\documentclass{article}

     \usepackage[final, nonatbib]{neurips_2020}

\usepackage[round]{natbib} 
\usepackage{placeins} 
\usepackage{adjustbox} 
\usepackage{amsmath}
\usepackage{xcolor} 
\usepackage{float} 
\usepackage{algorithmicx}
\usepackage[ruled]{algorithm}
\usepackage{algpseudocode}
\usepackage{subcaption}
\usepackage{bbm} 
\usepackage[colorlinks=false]{hyperref} 
\usepackage{wrapfig} 
\usepackage{verbatim} 

\usepackage[utf8]{inputenc} 
\usepackage[T1]{fontenc}    
\usepackage{url}            
\usepackage{booktabs}       
\usepackage{amsfonts}       
\usepackage{nicefrac}       
\usepackage{microtype}      

\title{Bayesian Meta-Learning for the Few-Shot Setting via Deep Kernels}

\author{%
  Massimiliano Patacchiola\\
  School of Informatics\\
  University of Edinburgh\\
  \texttt{mpatacch@ed.ac.uk} \\
  \And
  Jack Turner\\
  School of Informatics\\
  University of Edinburgh\\
  \texttt{jack.turner@ed.ac.uk} \\
  \And
  Elliot J. Crowley\\
  School of Engineering\\
  University of Edinburgh\\
  \texttt{elliot.j.crowley@ed.ac.uk} \\
  \And
  Michael O'Boyle\\
  School of Informatics\\
  University of Edinburgh\\
  \texttt{mob@inf.ed.ac.uk} \\
  \And
  Amos Storkey\\
  School of Informatics\\
  University of Edinburgh\\
  \texttt{a.storkey@ed.ac.uk} \\
}

\begin{document}

\maketitle

\begin{abstract}
    Recently, different machine learning methods have been introduced to tackle the challenging few-shot learning scenario that is, learning from a small labeled dataset related to a specific task. Common approaches have taken the form of meta-learning: learning to learn on the new problem given the old. Following the recognition that meta-learning is implementing learning in a multi-level model, we present a Bayesian treatment for the meta-learning inner loop through the use of deep kernels. As a result we can learn a kernel that transfers to new tasks; we call this~\emph{Deep Kernel Transfer~(DKT)}. This approach has many advantages: is straightforward to implement as a single optimizer, provides uncertainty quantification, and does not require estimation of task-specific parameters. We empirically demonstrate that DKT outperforms several state-of-the-art algorithms in few-shot classification, and is the state of the art for cross-domain adaptation and regression. We conclude that complex meta-learning routines can be replaced by a simpler Bayesian model without loss of accuracy.
\end{abstract}

\section{Introduction}
One of the key differences between state-of-the-art machine learning methods, such as deep learning \citep{lecun2015deep, schmidhuber2015deep}, and human learning is that the former needs a large amount of data in order to find relevant patterns across samples, whereas the latter acquires rich structural information from a handful of examples. Moreover, deep learning methods struggle in providing a measure of uncertainty, which is a crucial requirement when dealing with scarce data, whereas humans can effectively weigh up different alternatives given limited evidence. In this regard, some authors have suggested that the human ability for few-shot inductive reasoning could derive from a Bayesian inference mechanism~\citep{steyvers2006probabilistic, tenenbaum2011grow}. Accordingly, we argue that the natural interpretation of meta-learning as implementing learning in a hierarchical model, leads to a Bayesian equivalent through the use of deep kernel methods.

Deep kernels combine neural networks with kernels to provide scalable and expressive closed-form covariance functions~\citep{hinton2008using, wilson2016deep}. 
If one has a large number of small but related tasks, as in few-shot learning, it is possible to define a common prior that induces knowledge transfer. This prior can be a deep kernel with parameters shared across tasks, so that given a new unseen task it is possible to effectively estimate the posterior distribution over a query set conditioned on a small support set. In a meta-learning framework \citep{hospedales2020meta} this corresponds to a Bayesian treatment for the inner loop cycle.
This is our proposed approach, which we refer to as deep kernel learning with transfer, or \emph{Deep Kernel Transfer~(DKT)} for short. 

We derive two versions of DKT for both the regression and the classification setting, comparing it against recent methods on a standardized benchmark environment; the code is released with an open-source license\footnote{\url{https://github.com/BayesWatch/deep-kernel-transfer}}.
DKT has several advantages over other few-shot methods, which can be summarized as follows:

\begin{enumerate}
    \item \emph{Simplicity and efficiency}: it does not require any complex meta-learning optimization routines, it is straightforward to implement as a single optimizer as the inner loop is replaced by an analytic marginal likelihood computation, and it is efficient in the low-data regime.
    \item \emph{Flexibility}: it can be used in a variety of settings such as regression, cross-domain and within-domain classification, with state-of-the-art performance.
    \item \emph{Robustness}: it provides a measure of uncertainty with respect to new instances, that is crucial for a decision maker in the few-shot setting.
\end{enumerate}

\emph{Main contributions:} (i) a novel approach to deal with the few-shot learning problem through the use of deep kernels, (ii) an effective Bayesian treatment for the meta-learning inner-loop, and (iii) empirical evidence that complex meta-learning routines for few-shot learning can be replaced by a simpler hierarchical Bayesian model without loss of accuracy.

\subsection{Motivation}

The Bayesian meta-learning approach to the few-shot setting has predominantly followed the route of hierarchical modeling and multi-task learning~\citep{finn2018probabilistic, gordon2019meta, yoon2018bayesian}. The underlying directed graphical model distinguishes between a set of shared parameters $\boldsymbol{\theta}$, common to all tasks, and a set of $N$ task-specific parameters $\boldsymbol{\rho}_t$. Given a train dataset of tasks $\mathcal{D}=\{\mathcal{T}_t\}_{t=1}^{N}$, each one containing input-output pairs $\mathcal{T}=\{(x_l, y_l)\}_{l=1}^{L}$, and given a test point $x_{\ast}$ from a new task $\mathcal{T}_{\ast}$, learning consists of finding an estimate of $\boldsymbol{\theta}$, forming the posterior distribution over the task-specific parameters $p(\boldsymbol{\rho}_t | x_{\ast}, \mathcal{D}, \boldsymbol{\theta})$, and then computing the posterior predictive distribution $p(y_{\ast} | x_{\ast}, \boldsymbol{\theta})$. This approach is principled from a probabilistic perspective, but is problematic, as it requires managing two levels of inference via amortized distributions or sampling, often requiring cumbersome architectures. 

In recent differentiable meta-learning methods, the two sets of parameters are learned by maximum likelihood estimation, by iteratively updating $\boldsymbol{\theta}$ in an outer loop, and $\boldsymbol{\rho}_t$ in a inner loop \citep{finn2017model}. This case has various issues, since learning is destabilized by the joint optimization of two sets of parameters, and by the need to estimate higher-order derivatives (gradient of the gradient) for updating the weights \citep{antoniou2019train}.

To avoid these drawbacks we propose a simpler solution, that is marginalizing $\boldsymbol{\rho}_t$ over the data of a specific task. This marginalization is analytic and leads to a closed form marginal likelihood, which measures the~\emph{expectedness} of the data under the given set of parameters. By finding the parameters of a deep kernel we can maximize the marginal likelihood.
Following our approach there is no need to estimate the posterior distribution over the task-specific parameters, meaning that it is possible to directly compute the posterior predictive distribution, skipping an intermediate inference step.  We argue that this approach can be very effective in the few-shot setting, significantly reducing the complexity of the model with respect to meta-learning approaches, while retaining the advantages of Bayesian methods (e.g. uncertainty estimation) with state-of-the-art performances.

\section{Background}

\subsection{Few-shot Learning}\label{ssec_fewshot_learning}

The terminology describing the few-shot learning setup is dispersive due to the colliding definitions used in the literature; the reader is invited to see~\citet{chen2019closerfewshot} for a comparison. Here, we use the nomenclature derived from the meta-learning literature which is the most prevalent at time of writing. Let $\mathcal{S} = \{ (x_{l}, y_{l}) \}_{l=1}^L$ be a \emph{support-set} containing input-output pairs, with $L$ equal to one (1-shot) or five (5-shot), and  $\mathcal{Q} = \{ (x_{m}, y_{m}) \}_{m=1}^M$ be a \emph{query-set} (sometimes referred to in the literature as a target-set), with $M$ typically one order of magnitude greater than $L$. For ease of notation the support and query sets are grouped in a \emph{task} $\mathcal{T} = \{\mathcal{S}, \mathcal{Q}\}$, with the dataset $\mathcal{D} = \{ \mathcal{T}_{t} \}_{t=1}^{N}$ defined as a collection of such tasks. Models are trained on random tasks sampled from $\mathcal{D}$, then given a new task $\mathcal{T}_{\ast} = \{\mathcal{S}_{\ast}, \mathcal{Q}_{\ast} \}$ sampled from a test set, the objective is to condition the model on the samples of the support $\mathcal{S}_{\ast}$ to estimate the membership of the samples in the query set $\mathcal{Q}_{\ast}$. In the most common scenario, training, validation and test datasets each consist of distinct tasks sampled from the same overall distribution over tasks. Note that the target value $y$ can be a continuous value (regression) or a discrete one (classification), though most previous work has focused on classification. We also consider the \emph{cross-domain} scenario, where the test tasks are sampled from a different distribution over tasks than the training tasks; this is potentially more representative of many real-world scenarios.

\subsection{Kernels}

Given two input instances $x$ and $x^{\prime}$ and a function $f(\cdot)$, the kernel $k(x, x^{\prime})$ is a covariance function that expresses how the correlation of the outputs at two points depends on the relationship between their two locations in input space
\begin{equation}
    k(x,x^{\prime}) = \text{cov}(f(x),  f(x^{\prime})).
\end{equation}
The simplest kernel has a linear expression $k_\text{\tiny{LIN}}(x, x^{\prime}) = v \langle x, x^{\prime}\rangle$, where $\langle \cdot \rangle$ denotes an inner product, and $v$ is a variance hyperparameter. The use of a linear kernel is computationally convenient and it induces a form of Bayesian linear regression, however this is often too simplistic. For this reason, a variety of other kernels has been proposed in the literature: Radial Basis Function kernel (RBF), Mat\'ern kernel, Cosine Similarity kernel (CosSim), and the spectral mixture kernel \citep{wilson2013gaussian}. Details about the kernels used in this work are given in Appendix~\ref{sec:appendix_kernels}.

In deep kernel learning~\citep{hinton2008using, wilson2016deep} an input vector $\mathbf{x}$ is mapped to a latent vector $\mathbf{h}$ through a non-linear function $\mathcal{F}_{\phi}(\mathbf{x}) \rightarrow \mathbf{h}$ (e.g.\ a neural network) parameterized by a set of weights $\boldsymbol{\phi}$. The embedding is defined such that the dimensionality of the input is significantly reduced, meaning that if $\mathbf{x} \in \mathbb{R}^{J}$ and $\mathbf{h} \in \mathbb{R}^{K}$ then $J \gg K$. Once the input has been encoded in $\mathbf{h}$ the latent vector is passed to a kernel. When the inputs are images a common choice for $\mathcal{F}_{\phi}$ is a Convolutional Neural Network (CNN). Specifically we construct a kernel
\begin{equation}
    k(\mathbf{x}, \mathbf{x}^{\prime} | \boldsymbol{\theta},\boldsymbol{\phi})
    =
    k^\prime(\mathcal{F}_{\phi}(\mathbf{x}), \mathcal{F}_{\phi}(\mathbf{x}^{\prime}) |\boldsymbol{\theta})
\end{equation}
from some latent space kernel $k^\prime$ with hyperparameters $\boldsymbol{\theta}$ by passing the inputs through the non-linear function $\mathcal{F}_{\phi}$. Then the hyperparameters $\boldsymbol{\theta}$ and the parameters of the model $\boldsymbol{\phi}$ are jointly learned by maximizing the log marginal likelihood, backpropagating the error.

\section{Description of the method}
\label{sec:method}

Let us start from the interpretation of meta-learning as a hierarchical model \citep{finn2018probabilistic, grant2018recasting}, considering a set of task-common parameters in the upper hierarchy (optimized in an outer loop), along with a process for determining task-specific parameters in the lower hierarchy (optimized in an inner loop). For example, in MAML~\citep{finn2017model}, the outer-parameters are the common neural network initialization weights and the inner-parameters are the final network weights, with prior implicitly defined by the probability that a particular parameterization can be reached in a few gradient steps from the initial parameters. 
Both outer and inner loops are obtained end-to-end, by differentiating through the inner loop to obtain derivatives for the outer loop parameters. This causes well known instability problems \citep{antoniou2019train}.

\begin{algorithm}[H]
\small
\caption{Deep Kernel Transfer (DKT) in the few-shot setting, train and test functions.}
\label{alg_overview}
\textbf{Require:}  $\mathcal{D} = \{\mathcal{T}_n\}_{n=1}^{N}$ train dataset and $\mathcal{T}_{*}=\{\mathcal{S}_{*}, \mathcal{Q}_{*}\}$ test task. \\
\textbf{Require:}  $\boldsymbol{\hat{\theta}}$ kernel hyperparameters and $\boldsymbol{\hat{\phi}}$ neural network weights. \Comment{Randomly initialized} \\
\textbf{Require:}  $\alpha$, $\beta$: step size hyperparameters for the optimizers.
\begin{algorithmic}[1]
\vspace{0.1cm} 
\Function{Train}{$\mathcal{D}$, $\alpha$, $\beta$, $\boldsymbol{\hat{\theta}}$, $\boldsymbol{\hat{\phi}}$}
\While{not done}
    \State Sample task $\mathcal{T}=\{\mathcal{S}, \mathcal{Q}\} \sim \mathcal{D}$
        \State Assign $\mathcal{T}^{x} \leftarrow \forall x \in \mathcal{S} \cup \mathcal{Q}$ \ and \ $\mathcal{T}^{y} \leftarrow \forall y \in \mathcal{S} \cup \mathcal{Q}$
        \State $\mathcal{L}=-\log p(\mathcal{T}^{y} | \mathcal{T}^{x}, \boldsymbol{\hat{\theta}}, \boldsymbol{\hat{\phi}})$\Comment{see Eq. \eqref{eq:marglik}}
        \State Update $\boldsymbol{\hat{\theta}} \leftarrow \boldsymbol{\hat{\theta}} - \alpha \nabla_{\hat{\theta}} \mathcal{L}$ \ and \ $\boldsymbol{\hat{\phi}} \leftarrow \boldsymbol{\hat{\phi}} - \beta \nabla_{\hat{\phi}} \mathcal{L}$
\EndWhile
\State \Return $\boldsymbol{\hat{\theta}}$, $\boldsymbol{\hat{\phi}}$
\EndFunction
\vspace{0.1cm} 
\Function{Test}{$\mathcal{T}_{*}$, $\boldsymbol{\hat{\theta}}$, $\boldsymbol{\hat{\phi}}$}
        \State Assign $\mathcal{T}^{x}_{*} \leftarrow \forall x \in \mathcal{S}_{*}$, \ $\mathcal{T}^{y}_{*} \leftarrow \forall y \in \mathcal{S}_{*}$, \ and \ $x_{*} \leftarrow x \in \mathcal{Q}_{*}$
        \State \Return $p(y_{\ast} | x_{\ast}, \mathcal{T}_{\ast}^x,\mathcal{T}_{\ast}^y, \boldsymbol{\hat{\theta}}, \boldsymbol{\hat{\phi}})$ \Comment{see Eq. \eqref{eq_mlii_approximation}}
\EndFunction
\Statex
\end{algorithmic}
  \vspace{-0.2cm}%
\end{algorithm}

Our proposal is to replace the inner loop with a Bayesian integral, while still optimizing for the parameters. This is commonly called a \emph{maximum likelihood type II (ML-II)} approach. Specifically, we learn a set of parameters and hyperparameters of a deep kernel (outer-loop) that maximize a marginal likelihood across all tasks. The marginalization of this likelihood integrates out over each of the task-specific parameters for each task using a Gaussian process approach, replacing the inner loop model with a kernel.

Let all the input data (support and query) for task $t$ be denoted by $\mathcal{T}^x_{t}$ and the target data be $\mathcal{T}^y_{t}$. Let $\mathcal{D}^x$ and $\mathcal{D}^y$ denote the respective collections of these datasets over all tasks; this data is hierarchically grouped by task. The marginal-likelihood of a Bayesian hierarchical model, conditioned on task-common hyperparameters $\boldsymbol{\hat{\theta}}$ and other task-common parameters $\boldsymbol{\hat{\phi}}$ (e.g. neural network weights) would take the form
\begin{equation}
P(\mathcal{D}^y|\mathcal{D}^x,\boldsymbol{\hat{\theta}}, \boldsymbol{\hat{\phi}})=\prod_t P(\mathcal{T}^y_t|\mathcal{T}^x_t,\boldsymbol{\hat{\theta}}, \boldsymbol{\hat{\phi}}),
\label{eq:marglik}
\end{equation}
where $P(\mathcal{T}^y_t|\mathcal{T}^x_t,\boldsymbol{\hat{\theta}}, \boldsymbol{\hat{\phi}})$ is a marginalization over each set of task-specific parameters. Let these task-specific parameters for task $t$ be denoted by $\boldsymbol{\rho}_t$, then
\begin{equation}
P(\mathcal{T}^y_t|\mathcal{T}^x_t,\boldsymbol{\theta}, \boldsymbol{\phi})= \int \prod_k P(y_k|x_k,\boldsymbol{\theta}, \boldsymbol{\phi},\boldsymbol{\rho}_t) d\boldsymbol{\rho}_t,
\label{eq:intout}
\end{equation}
where $k$ enumerates elements of $x_k\in \mathcal{T}^x_t$, and corresponding elements $y_k \in \mathcal{T}^y_t$. 
In typical meta-learning, the task-specific integral \eqref{eq:intout} would be replaced by an inner-loop optimizer for the task-specific objective (and the parameters of that optimizer); any additional cross-task parameters $\boldsymbol{\theta}$, $\boldsymbol{\phi}$ would be optimized in the outer loop. Instead, we do a full integral of the task specific parameters, and optimize only for the cross-task parameters $\boldsymbol{\theta}$, $\boldsymbol{\phi}$. We do that implicitly rather than explicitly by using a Gaussian process model for $P(\mathcal{T}^y_t|\mathcal{T}^x_t,\boldsymbol{\theta})$, which is the outcome of an analytic integral of Equation~\ref{eq:intout} for many model classes \citep{rasmussen2006gaussian}.
Predictions of the value $y_*$ for a new point $x_*$ given a small set of exemplars $\mathcal{T}_{t_*}^x,\mathcal{T}_{t_*}^y$ for a new task $t_*$ can be made using the predictive distribution
\begin{equation}\label{eq_mlii_approximation}
p(y_{\ast} | x_{\ast}, \mathcal{T}_{t_\ast}^x,\mathcal{T}_{t_\ast}^y) \approx p(y_{\ast} | x_{\ast}, \mathcal{T}_{t_\ast}^x,\mathcal{T}_{t_\ast}^y, \boldsymbol{\hat{\theta}}, \boldsymbol{\hat{\phi}}).
\end{equation}

Our claim is that, though the number of data points for each task is potentially small, the total number of points over all tasks contributing to the marginal likelihood~\eqref{eq:marglik} is sufficiently large to make ML-II appropriate for finding a set of shared weights and parameters without underfitting or overfitting. Those parameters provide a model with good generalization capability to new unseen tasks, removing the need for inferring the task-specific parameters $\boldsymbol{\rho}_t$. Results in Section~\ref{sec:experiments} indicate that our proposal is competitive with much more complicated meta-learning methods. Note that this approach differs from direct deep kernel learning, where the marginalization is over all data; this would ignore the task distinctions which is vital given the hierarchical mode (see experimental comparison in Section~\ref{ssec:experiments_regression}). The problem also differs from multitask learning where the tasks share the same input values.

For stochastic gradient training, at each iteration, a task $\mathcal{T}=\{\mathcal{S}, \mathcal{Q}\}$ is sampled from $\mathcal{D}$, then the log marginal likelihood, that is the logarithm of \eqref{eq:marglik}, is estimated over $\mathcal{S} \cup \mathcal{Q}$ (assuming $y \in Q$ to be observed) and the parameters of the kernel are updated via a gradient step on the marginal likelihood objective for that task. This procedure allows us to find a kernel that can represent the task in its entirety over both support and query sets. At test time, given a new task $\mathcal{T}_{*}=\{\mathcal{S}_{*}, \mathcal{Q}_{*}\}$ the prediction on the query set $\mathcal{Q}_{*}$ is made via conditioning on the support set $\mathcal{S}_{*}$, using the parameters that have been learned at training time. Pseudocode is given in Algorithm~\ref{alg_overview}.

\subsection{Regression}
We want to find a closed form expression of \eqref{eq:marglik} for the regression case. Assume we are interested in a continuous output $y_{\ast}$ generated by a clean signal $f_{\ast}(x_{\ast})$ corrupted by homoscedastic Gaussian noise $\epsilon$ with variance $\sigma^{2}$. We are interested in the joint distribution of the observed outputs and the function values at test location. For ease of notation, let us define $\mathbf{k_{\ast}}= k(x_{\ast},\mathbf{x})$ to denote the $N$-dimensional vector of covariances between $x_{\ast}$ and the $N$ training points in $\mathcal{D}$. Similarly, let us write $k_{\ast \ast} = k(x_{\ast},x_{\ast})$ for the variance of $x_{\ast}$, and $\mathbf{K}$ to identify the covariance matrix on the training inputs in $\mathcal{D}$. The predictive distribution $p(y_{\ast} | x_{\ast}, \mathcal{D})$ is obtained by Bayes' rule, and given the conjugacy of the prior, this is a Gaussian with mean and covariance specified as
\begin{subequations}\label{eq_posterior_function_view}
\begin{alignat}{2}
  \mathbb{E}[ f_{\ast} ] &= 
 \mathbf{k_{\ast}}^{\top} (\mathbf{K} + \sigma^2 \mathbf{I})^{-1} \mathbf{y},\\
 \text{cov}(f_{\ast}) &= k_{\ast \ast} - \mathbf{k_{\ast}}^{\top} (\mathbf{K} + \sigma^2 \mathbf{I})^{-1} 
 \mathbf{k_{\ast}}.
 \end{alignat}
\end{subequations}
Note that \eqref{eq_posterior_function_view} defines a distribution over functions, which assumes that the collected values at any finite set of points have a joint Gaussian distribution \citep{rasmussen2006gaussian}. Hereon, we absorb the noise $\sigma^2 \mathbf{I}$ into the covariance matrix $\mathbf{K}$ and treat it as part of a vector of learnable parameters $\boldsymbol{\theta}$, that also include the hyperparameters of the kernel (e.g. variance of a linear kernel). 

Let us collect all the target data items for task $t$ into vector $\mathbf{y}_t$, and denote the kernel between all task inputs by $K_t$. It follows that the marginal likelihood of Equation~\eqref{eq:marglik} can be rewritten as

\begin{equation}
{\small
\log P(\mathcal{D}^y|\mathcal{D}^x,\boldsymbol{\hat{\theta}}, \boldsymbol{\hat{\phi}})
= \sum_t -
\underbrace{\frac{1}{2} \mathbf{y}_t^{\top} [K_t(\boldsymbol{\hat{\theta}}, \boldsymbol{\hat{\phi}})]^{-1} \mathbf{y}_t}_{\text{data-fit}}
-
\underbrace{\frac{1}{2} \log |K_t(\boldsymbol{\hat{\theta}}, \boldsymbol{\hat{\phi}})|}_{\text{penalty}}
+c,
\label{eq_log_marginal_likelihood_tractable}
}
\end{equation}
where $c$ is a constant. The parameters are estimated via ML-II maximizing~\eqref{eq_log_marginal_likelihood_tractable} via gradient ascent. In practice we use a stochastic gradient ascent with each batch containing the data for a single task.

\subsection{Classification}

A Bayesian treatment for the classification case does not come without problems, since a non-Gaussian likelihood breaks the conjugacy. For instance, in the case of binary classification the Bernoulli likelihood induces an intractable marginalization of the evidence and therefore it is not possible to estimate the posterior in a closed form. Common approaches to deal with this issue (e.g. MCMC or variational methods), incur a significant computational cost for few-shot learning: for each new task, the posterior is estimated by approximation or sampling, introducing an inner loop that increases the time complexity from constant $\mathcal{O}(1)$ to linear $\mathcal{O}(K)$, with $K$ being the number of inner cycles. An alternative solution would be to treat the classification problem as if it were a regression one, therefore reverting to analytical expressions for both the evidence and the posterior. In the literature this has been called~\emph{label regression (LR)}~\citep{kuss2006gaussian} or \emph{least-squares classification (LSC)} \citep{rifkin2004defense, rasmussen2006gaussian}. Experimentally, LR and LSC tend to be more effective than other approaches in both binary~\citep{kuss2006gaussian} and multi-class~\citep{rifkin2004defense} settings. Here, we derive a classifier based on LR which is computationally cheap and straightforward to implement.

Let us define a binary classification setting, with the class being a Bernoulli random variable $c \in \{ 0,1 \}$. The model is trained as a regressor with a target $y_{+}=1$ to denote the case $c=1$, and $y_{-}=-1$ to denote the case $c=0$. Even though $y \in \{-1,1\}$ there is no guarantee that $f(x) \in [y_{-}, y_{+}]$. Predictions are made by computing the predictive mean and passing it through a sigmoid function, inducing a probabilistic interpretation. Note that it is still possible to use ML-II to make point estimates of $\boldsymbol{\theta}$ and $\boldsymbol{\phi}$. When generalizing to a multi-label task we apply the \emph{one-versus-rest} scheme where $C$ binary classifiers are used to classify each class against all the rest. The log marginal likelihood, that is the logarithm of Equation \eqref{eq:marglik}, is replaced by the sum of the marginals for each one of the $C$ individual class outputs $\mathbf{y}_c$, as
\begin{equation}\label{eq_classification_log_marginal_likelihood}
    \log p(\mathbf{y} | \mathbf{x}, \boldsymbol{\hat{\theta}}, \boldsymbol{\hat{\phi}})=
\sum_{c=1}^{C}
\log p(\mathbf{y}_c | \mathbf{x}, \boldsymbol{\hat{\theta}}, \boldsymbol{\hat{\phi}}).
\end{equation}
Given a new input $x_{\ast}$ and the $C$ outputs of all the binary classifiers, a decision is made by selecting the output with the highest probability $c_{\ast} = \mathrm{argmax}_{c} \big( \sigma(m_{c}(x_{\ast})) \big),$ where $m(x)$ is the predictive mean, $\sigma(\cdot)$ the sigmoid function, and $c_{\ast} \in \{1, ..., C\}$.

\section{Related Work}\label{sec_related_work}
There is a wealth of literature on feature transfer~\citep{pan2009survey}. As a baseline for few-shot learning, the standard procedure consists of two phases: pre-training and fine-tuning. During pre-training, a network and classifier are trained on examples for the base classes. When fine-tuning, the network parameters are fixed and a new classifier is trained on the novel classes. This approach has its limitations; part of the model has to be trained from scratch for each new task, and often overfits.~\citet{chen2019closerfewshot} extend this by proposing the use of cosine distance between examples (called Baseline$++$). However, this still relies on the assumption that a fixed fine-tuning protocol will balance the bias-variance tradeoff correctly for every task. 

Alternatively, one can compare new examples in a learned metric space. Matching Networks (MatchingNets,~\citealp{vinyals2016matching}) use a softmax over cosine distances as an attention mechanism, and an LSTM to encode the input in the context of the support set, considered as a sequence. Prototypical Networks (ProtoNets,~\citealp{snell2017prototypical}) are based on learning a metric space in which classification is performed by computing distances to prototypes, where each prototype is the mean vector of the embedded support points belonging to its class.
Relation Networks (RelationNets,~\citealp{sung2018learning}) use an embedding module to generate representations of the query images that are compared by a relation module with the support set, to identify matching categories.

Meta-learning~\citep{bengio1992optimization, schmidhuber1992learning, hospedales2020meta} methods have become very popular for few-shot learning tasks. MAML~\citep{finn2017model} has been proposed as a way to meta-learn the parameters of a model over many tasks, so that the initial parameters are a good starting point from which to adapt to a new task. MAML has provided inspiration for numerous meta-learning approaches~\citep{antoniou2019train, rajeswaran2019meta}.

In several works, MAML has been interpreted as a Bayesian hierarchical model~\citep{finn2018probabilistic, grant2018recasting, jerfel2019reconciling}. Bayesian MAML \citep{yoon2018bayesian} combines efficient gradient-based meta-learning with nonparametric variational inference, while keeping an application-agnostic approach. \citet{gordon2019meta} have recently proposed an amortization network---VERSA---that takes few-shot learning datasets as inputs, and outputs a distribution over task-specific parameters which can be used to meta-learn probabilistic inference for prediction. \cite{xu2019metafun} have used conditional neural processes with an encoder-decoder architecture to project labeled data into an infinite-dimensional functional representation.

For the regression case \cite{harrison2018meta} have proposed a method named ALPaCA, which uses a dataset of sample functions to learn a domain-specific encoding and a prior over weights. \citet{tossou2019adaptive} have presented a variant of kernel learning for Gaussian Processes called Adaptive Deep Kernel Learning (ADKL), which finds a kernel for each task using a task encoder network. The difference between our method and ADKL is that we do not need an additional module for task encoding as we can rely on a single set of shared general-purpose hyperparameters.

\section{Experiments}
\label{sec:experiments}

\begin{figure}[t!]
    \begin{subfigure}[t]{0.36\textwidth}
        \centering
        \includegraphics[width=0.98\textwidth, trim={0.0cm 0.0cm 0.0cm 0.0cm}, clip]{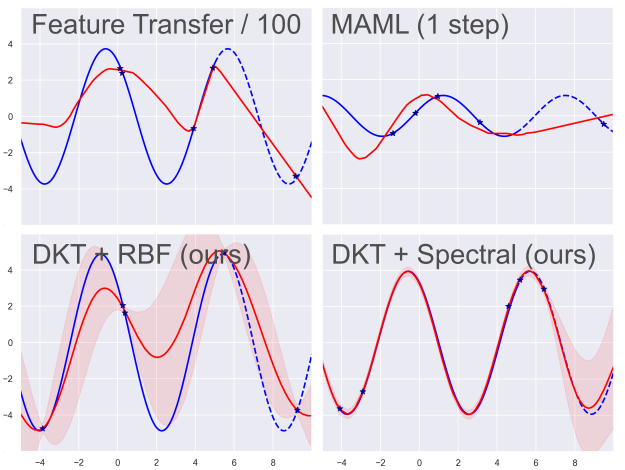}
        \caption{Qualitative comparison}
        \label{fig:results-regression-comparison}
    \end{subfigure}
    \begin{subfigure}[t]{0.64\textwidth}
        \includegraphics[width=0.98\textwidth, trim={0.0cm 0.0cm 0.0cm 0.0cm}, clip]{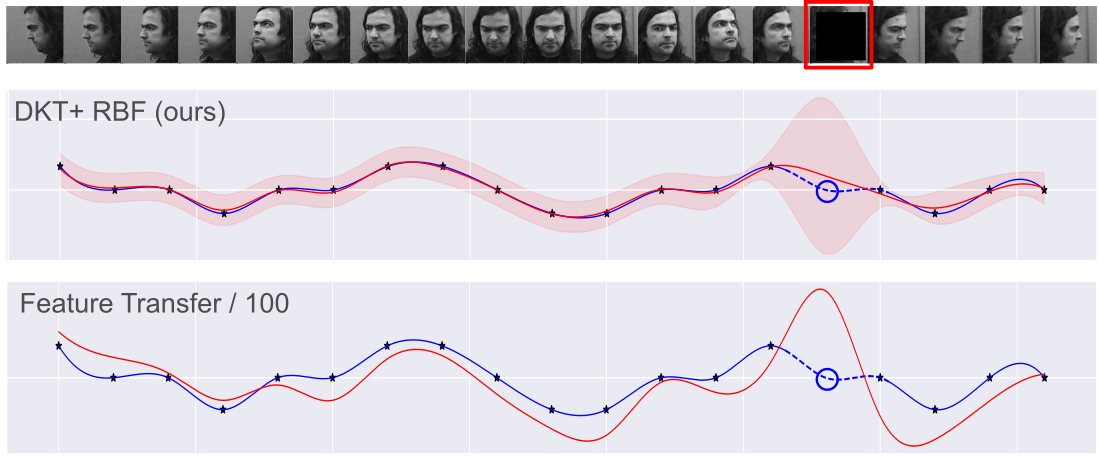}
        \caption{Uncertainty estimation}
        \label{fig:results-regression-uncertainty}
    \end{subfigure}
    \caption{(\subref{fig:results-regression-comparison})~Comparison between different methods for unknown function approximation (out-of-range, 5 support points). DKT better fits (red line) the true function (solid blue) and the out-of-bound portion never seen at training time (dashed blue). Uncertainty (red shadow) increases in low-confidence regions. (\subref{fig:results-regression-uncertainty})~Uncertainty estimation for an outlier (Cutout noise, red frame) in the head trajectory estimation from images. DKT is able to estimate a mean value (red line) close to the true value (blue circle) showing large variance. Feature transfer performs poorly at the same location. }
    \label{fig_plot_regression_results}
\end{figure}

In the few-shot setting a fair comparison between methods is often obfuscated by substantial differences in the implementation details of each algorithm. \citet{chen2019closerfewshot} have recently released an open-source benchmark to allow for a fair comparison between methods. We integrated our algorithm into this framework using PyTorch and GPyTorch~\citep{gardner2018gpytorch}. In all experiments the proposed method is marked as~\emph{DKT}. Training details are reported in Appendix~\ref{appendix:training_details}.

\subsection{Regression}\label{ssec:experiments_regression}

\begin{wraptable}{R}{0.45\textwidth}
\caption{Average Mean-Squared Error (MSE) and standard deviation (three runs) on few-shot regression for periodic functions (top) and head pose trajectory (bottom), using 10 samples for train, and 5 for test. Same domain is marked as \emph{in-range}, extended unseen domain as \emph{out-of-range}. Lowest error in bold. $^\ast$Results reported by \cite{tossou2019adaptive}.}
\begin{adjustbox}{width=0.45\columnwidth,center}
\begin{tabular}{lcc}
\hline
\small{\textbf{Method}} & \textbf{in-range} & \textbf{out-of-range}\\ 
\hline
\multicolumn{3}{c}{\small{\underline{Periodic functions}}}\\
\small{\textbf{ADKL}} \citep{tossou2019adaptive}$^\ast$ & \small{0.14} & \small{--}\\
\small{\textbf{R2-D2}} \citep{bertinetto2019meta}$^\ast$ & \small{0.46} & \small{--}\\
\small{\textbf{ALPaCA}} \citep{harrison2018meta} & \small{0.14 $\pm$ 0.09} & \small{5.92 $\pm$ 0.11}\\
\small{\textbf{Feature Transfer/1}}
& \small{2.94 $\pm$ 0.16} & \small{6.13 $\pm$ 0.76}\\
\small{\textbf{Feature Transfer/100}}
& \small{2.67 $\pm$ 0.15} & \small{6.94 $\pm$ 0.97}\\
\small{\textbf{MAML (1 step)}} & \small{2.76 $\pm$ 0.06} & \small{8.45 $\pm$ 0.25}\\
\small{\textbf{DKBaseline $+$ RBF}} & \small{2.85 $\pm$ 1.14} & \small{3.65 $\pm$ 1.63} \\
\small{\textbf{DKBaseline $+$ Spectral}} & \small{2.08 $\pm$ 2.31} & \small{4.11 $\pm$ 1.92} \\ 
\small{\textbf{DKT $+$ RBF}} (ours) & \small{1.38 $\pm$ 0.03} & \small{2.61 $\pm$ 0.16}\\
\small{\textbf{DKT $+$ Spectral}} (ours) & \small{\textbf{0.08 $\pm$ 0.06}} & \small{\textbf{0.10 $\pm$ 0.06}}\\
\hline
\multicolumn{3}{c}{\small{\underline{Head pose trajectory}}}\\
\small{\textbf{Feature Transfer/1}}     & \small{0.25 $\pm$ 0.04}   & \small{0.20 $\pm$ 0.01}\\
\small{\textbf{Feature Transfer/100}}   & \small{0.22 $\pm$ 0.03}   & \small{0.18 $\pm$ 0.01}\\
\small{\textbf{MAML (1 step)}}           & \small{0.21 $\pm$ 0.01}         & \small{0.18 $\pm$ 0.02}\\
\small{\textbf{DKT $+$ RBF}} (ours) & \small{0.12 $\pm$ 0.04}  & \small{0.14 $\pm$ 0.03}\\
\small{\textbf{DKT $+$ Spectral}} (ours) & \small{\textbf{0.10 $\pm$ 0.01}}   & \small{\textbf{0.11 $\pm$ 0.02}}\\
\hline
\end{tabular}
\end{adjustbox}
\label{tab_regression_mse}
\end{wraptable}

We consider two tasks: amplitude prediction for unknown periodic functions, and head pose trajectory estimation from images. The former was treated as a few-shot regression problem by~\citet{finn2017model} to motivate MAML: support and query scalars are uniformly sampled from a periodic wave with amplitude $\in [0.1, 5.0]$, phase $\in [0, \pi]$, and range $\in [-5.0, 5.0]$, and Gaussian noise ($\mu=0,~\sigma=0.1$). The training set is composed of 5 support and 5 query points, and the test set of 5 support and 200 query points. We first test~\emph{in-range}: the same domain as the training set as in~\citet{finn2017model}. We also consider a more challenging~\emph{out-of-range} regression, with test points drawn from an extended domain $[-5.0, 10.0]$ where portions from the range $[5.0, 10.0]$ have~\emph{not been seen} at training time.

For head pose regression, we used the Queen Mary University of London multiview face dataset (QMUL,~\citealp{gong1996investigation}), it comprises of grayscale face images of 37 people (32 train, 5 test). There are 133 facial images per person, covering a viewsphere of $\pm 90^{\circ}$ in yaw and $\pm 30^{\circ}$ in tilt at $10^{\circ}$ increment. Each task consists of randomly sampled trajectories taken from this discrete manifold, where~\emph{in-range} includes the full manifold and~\emph{out-of-range} allows training only on the leftmost 10 angles, and testing on the full manifold; the goal is to predict tilt.
For the periodic function prediction experiment, we compare our approach against feature transfer and MAML~\citep{finn2017model}. Moreover we report the results of ADKL~\citep{tossou2019adaptive}, R2-D2~\citep{bertinetto2019meta}, and ALPaCA~\citep{harrison2018meta} obtained on a similar task (as defined in \citealp{yoon2018bayesian}). To highlight the importance of kernel transfer, we add a baseline where a deep kernel is trained from scratch on the support points of every incoming task without transfer (DKBaseline), this correspond to standard deep kernel learning~\citep{wilson2016deep}.
Few methods have tackled few-shot regression from images, so in the head pose trajectory estimation we compare against feature transfer and MAML. As metric we use the average Mean-Squared Error (MSE) between predictions and true values. Additional details are reported in Appendix~\ref{appendix:training_details}.

Results for the regression experiments are summarized in Table~\ref{tab_regression_mse} and a qualitative comparison is provided in Figure~\ref{fig:results-regression-comparison} and in appendix. DKT obtains a lower MSE than feature transfer and MAML on both experiments. For unknown periodic function estimation, using a spectral kernel gives a large advantage over RBF, being more precise in both in-range and out-of-range (1.38 vs 0.08, and 2.61 vs 0.10 MSE). Uncertainty is correctly estimated in regions with low point density, and increases overall in the out-of-range region. Conversely, feature transfer severely underfits (1 step, 2.94 MSE) or overfits (100 step, 2.67), and was unable to model out-of-range points (6.13 and 6.94). MAML is effective in-range (2.76), but significantly worse out-of-range (8.45). ADKL, R2-D2, and ALPaCA (0.14, 0.46, 0.14) are better than DKT with an RBF kernel (1.38), but worse than DKT with a Spectral kernel (0.08). This indicates that the combination of an appropriate kernel with our method is more effective than an adaptive approach. The DKBaseline performs significantly worse than DKT in all conditions, confirming the necessity of using kernel transfer for few-shot problems.
Qualitative comparison in Figure~\ref{fig:results-regression-comparison} shows that both feature transfer and MAML are unable to fit the true function, especially out-of-range; additional samples are reported in Appendix \ref{sec:appendix_results_regression}. We observe similar results for head pose estimation, with DKT reporting lower MSE in all cases (Table~\ref{tab_regression_mse}). 
In Appendix~\ref{sec:appendix_results_regression} we also examine the latent spaces generated by the RBF and spectral kernel.

\textbf{Uncertainty quantification (regression)} In the low-data regime it is fundamental to account for uncertainty in the prediction; DKT is one of the few methods able to do it. 
To highlight the benefits of our method versus other approaches, we perform an experiment on~\emph{quantifying uncertainty}, sampling head pose trajectories and corrupting one input with Cutout~\citep{devries2017improved}, randomly covering $95\%$ of the image. Qualitative results are shown in Figure~\ref{fig:results-regression-uncertainty}. For the corrupted input, DKT predicts a value close to the true one, while giving a high level of uncertainty (red shadow). Feature transfer performs poorly, predicting an unrealistic pose.

\subsection{Classification}
We consider two challenging datasets: the Caltech-UCSD Birds (CUB-200,~\citealp{wah2011caltech}), and mini-ImageNet~\citep{ravi2017optimization}. All the experiments are 5-way (5 random classes) with 1 or 5-shot (1 or 5 samples per class in the support set). A total of 16 samples per class are provided for the query set. Additional details in Appendix~\ref{appendix:training_details}.
We compare the following kernels: linear, RBF, Mat\'ern, Polynomial, CosSim, and BNCosSim. Where BNCosSim is a variant of CosSim with features centered through BatchNorm (BN) statistics \citep{ioffe2015batch}, this has shown to improve performance \citep{wang2019simpleshot}.
We compare our approach to several state-of-the-art methods, such as MAML~\citep{finn2017model}, ProtoNets~\citep{snell2017prototypical}, MatchingNet~\citep{vinyals2016matching}, and RelationNet~\citep{sung2018learning}. We further compare against feature transfer, and Baseline$++$ from~\citet{chen2019closerfewshot}. All these methods have been trained from scratch with the same backbone and learning schedule. We additionally report the results for approaches with comparable training procedures and convolutional architectures \citep{mishra2018simple, ravi2017optimization, wang2019simpleshot} including recent hierarchical Bayesian methods \citep{gordon2019meta, grant2018recasting, jerfel2019reconciling}. We have excluded approaches that use deeper backbones or more sophisticated learning schedules \citep{antoniou2019learning, oreshkin2018tadam, qiao2018few, ye2018learning} so that the quality of the algorithms can be assessed separately from the power of the underlying discriminative model.

We report the results for the more challenging 1-shot case in Table \ref{tab:results_1shot_cub_cross} and \ref{tab:results_1shot_mini}, and the results for the 5-shot case in appendix. DKT achieves the highest accuracy in both CUB (63.37\%) and mini-ImageNet (49.73\%), performing better than any other approach including hierarchical Bayesian methods such as LLAMA (49.40\%) and VERSA (48.53\%). The best performance of first-order kernels (Table~\ref{tab_classification_kernel_comparison}, in appendix) is likely due to a low-curvature manifold induced by the neural network in the latent space, increasing the linear separability of data. Overall our results confirm the findings of~\citet{chen2019closerfewshot} regarding the effectiveness of cosine metrics, and those of~\citet{wang2019simpleshot} on the importance of feature normalization (Appendix~\ref{sec:appendix_results_classification} and \ref{sec:appendix_results_crossdomain}). In Table~\ref{tab_backbone_accuracy} (in appendix) we report results with a deeper backbone (ResNet-10, \citealt{he2016deep}), showing that DKT outperforms all other methods in 5-shot (85.64\%) with the second best result in 1-shot (72.27\%). The difference in performance between CosSim and BNCosSim is larger for the deeper backbone, indicating that centering the features is important when additional layers are added to the network.

\begin{table}[]
    \begin{minipage}{0.6\linewidth}
\caption{Average accuracy and standard deviation (percentage) over three runs (1-shot, 5-ways, Conv-4) on CUB and cross-domain classification (Omniglot to EMNIST and mini-Imagenet to CUB). Best results highlighted in bold.}
\begin{adjustbox}{width=1.0\columnwidth,center}
\begin{tabular}{lccc}
\hline
\textbf{\small{Method}} & \textbf{\small{CUB}} & \textbf{\small{Omni}}$\rightarrow$\small{\textbf{EMNIST}} & \textbf{\small{ImgNet}}$\rightarrow$\textbf{\small{CUB}} \\
\hline
\small{\textbf{Feature Transfer}} & 46.19 $\pm$ \small{0.64} & 64.22 $\pm$ \small{1.24} & 32.77 $\pm$ \small{0.35} \\
\small{\textbf{Baseline$++$}} \citep{chen2019closerfewshot} & 61.75 $\pm$ \small{0.95} & 56.84 $\pm$ \small{0.91} & 39.19 $\pm$ \small{0.12} \\
\small{\textbf{MatchingNet}} \citep{vinyals2016matching}  & 60.19 $\pm$ \small{1.02} & 75.01 $\pm$ \small{2.09} & 36.98 $\pm$ \small{0.06} \\
\small{\textbf{ProtoNet}} \citep{snell2017prototypical} & 52.52 $\pm$ \small{1.90} & 72.04 $\pm$ \small{0.82} & 33.27 $\pm$ \small{1.09} \\
\small{\textbf{MAML}} \citep{finn2017model} & 56.11 $\pm$ \small{0.69} & 72.68 $\pm$ \small{1.85} & 34.01 $\pm$ \small{1.25} \\
\small{\textbf{RelationNet}} \citep{sung2018learning} & 62.52 $\pm$ \small{0.34} & 75.62 $\pm$ \small{1.00} & 37.13 $\pm$ \small{0.20} \\
\hline
\small{\textbf{DKT + Linear}} (ours) & 60.23 $\pm$ \small{0.76} & \textbf{75.97 $\pm$ \small{0.70}} & 38.72 $\pm$ \small{0.42} \\
\small{\textbf{DKT + CosSim}} (ours) & \textbf{63.37 $\pm$ \small{0.19}} & 73.06 $\pm$ \small{2.36} & \textbf{40.22 $\pm$ \small{0.54}} \\
\small{\textbf{DKT + BNCosSim}} (ours) & 62.96 $\pm$ \small{0.62} & 75.40 $\pm$ \small{1.10} & 40.14 $\pm$ \small{0.18} \\
\hline
\label{tab:results_1shot_cub_cross}
\end{tabular}
\end{adjustbox}
    \end{minipage}
    \quad
    \begin{minipage}{0.37\linewidth}
\caption{Classification: 1-shot, 5-ways, Conv-4. Best results in bold.}
\begin{adjustbox}{width=1.0\columnwidth,center}
\begin{tabular}{lc}
\hline
\small{\textbf{Method}} & \textbf{\small{mini-ImageNet}}\\
\hline
\small{\textbf{ML-LSTM}} \citep{ravi2017optimization} & 43.44 $\pm$ \small{0.77} \\
\small{\textbf{SNAIL}} \citep{mishra2018simple} & 45.10\\
\small{\textbf{iMAML-HF}} \citep{rajeswaran2019meta} & 49.30 $\pm$ \small{1.88}\\
\small{\textbf{LLAMA}} \citep{grant2018recasting} & 49.40 $\pm$ \small{1.83}\\
\small{\textbf{VERSA}} \citep{gordon2019meta} & 48.53 $\pm$ \small{1.84}\\
\small{\textbf{Amortized VI}} \citep{gordon2019meta} & 44.13 $\pm$ \small{1.78} \\
\small{\textbf{Meta-Mixture}} \citep{jerfel2019reconciling} & 49.60 $\pm$ \small{1.50} \\
\small{\textbf{SimpleShot}} \citep{wang2019simpleshot} & 49.69 $\pm$ \small{0.19} \\
\small{\textbf{Feature Transfer}} & 39.51 $\pm$ \small{0.23} \\
\small{\textbf{Baseline$++$}} \citep{chen2019closerfewshot} & 47.15 $\pm$ \small{0.49} \\
\small{\textbf{MatchingNet}} \citep{vinyals2016matching} & 48.25 $\pm$ \small{0.65} \\
\small{\textbf{ProtoNet}} \citep{snell2017prototypical} & 44.19 $\pm$ \small{1.30} \\
\small{\textbf{MAML}}  \citep{finn2017model} & 45.39 $\pm$ \small{0.49} \\
\small{\textbf{RelationNet}} \citep{sung2018learning} & 48.76 $\pm$ \small{0.17} \\
\hline
\small{\textbf{DKT + CosSim}} (ours) & 48.64 $\pm$ \small{0.45} \\
\small{\textbf{DKT + BNCosSim}} (ours) & \textbf{49.73 $\pm$ \small{0.07}} \\
\hline
\label{tab:results_1shot_mini}
\end{tabular}
\end{adjustbox}
    \end{minipage}
\end{table}

\textbf{Uncertainty quantification (classification)} We provide results on model calibration on the CUB dataset. We followed the protocol of \cite{guo2017calibration} estimating the Expected Calibration Error (ECE), a scalar summary statistic (the lower the better). We first scaled each model output, calibrating the temperature by minimizing the NLL on logits/labels via LBFGS on 3000 tasks; then we estimated the ECE on the test set. The complete results for CUB 1-shot and 5-shot (percentage, average of three runs) are reported in Appendix~\ref{sec:appendix_results_classification}, Table~\ref{tab_uncertainty_ece}. In 1-shot DKT achieves one of the lowest ECE $2.6\%$ beating most of the competitors (only ProtoNet and MAML do better). In 5-shot our model achieves the second lowest ECE $1.1\%$ (ProtoNet does marginally better).

\subsection{Cross-domain classification}
The objective of cross-domain classification is to train a model on tasks sampled from one distribution, that then generalizes to tasks sampled from a different distribution. Specifically, we combine datasets so that the training split is drawn from one, and the validation and test split are taken from another. We experiment on mini-ImageNet$\rightarrow$CUB (train split from mini-ImageNet and val/test split from CUB) and Omniglot$\rightarrow$EMNIST. We compare our method to the previously-considered approaches, using identical settings for number of epochs and model selection strategy (see Appendix~\ref{appendix:training_details}).
Results for the 1-shot case are given in Table~\ref{tab:results_1shot_cub_cross}. DKT achieves the highest accuracy in most conditions. In Omniglot$\rightarrow$EMNIST, the best performance is achieved with a linear kernel (75.97\%). In mini-ImageNet$\rightarrow$CUB, DKT surpasses all the other methods obtaining the highest accuracy with CosSim (40.22\%) and BNCosSim (40.14\%). Note that most competing methods experience difficulties in this setting, as shown by their low accuracies and large standard deviations. A comparison of kernels shows that first order ones are more effective (see Appendix~\ref{sec:appendix_results_crossdomain}, Table~\ref{tab_crossdomain_kernel_comparison}).

\section{Conclusion}
In this work, we have introduced DKT, a highly flexible Bayesian model based on deep kernel learning. Compared with some other approaches in the literature, DKT performs better in regression and cross-domain classification while providing a measure of uncertainty. Based on the results, we argue that many complex meta-learning routines for few-shot learning can be replaced by a simple hierarchical Bayesian model without loss of accuracy. Future work could focus on exploiting the flexibility of the model in related settings, especially those merging continual and few-shot learning \citep{antoniou2020defining}, where DKT has the potential to thrive.

\FloatBarrier
\section*{Broader Impact}

The main motivation of this work has been to design a simple yet effective Bayesian method for dealing with the few-shot learning setting. The ability to learn from a reduced amount of data is crucial if we want to have systems that are able to deal with concrete real-world problems. Applications include (but are not limited to): classification and regression under constrained computational resources, medical diagnosis from small datasets, biometric identification from a handful of images, etc. 
Our method is one of the few which is able to provide a measure of uncertainty as a feedback for the decision maker. However, it is important to wisely choose the data on which the system is trained, since the low-data regime may be prone to bias more than the standard counterpart. If data is biased our method is not guaranteed to provide a correct estimation; this could harm the final users and should be carefully taken into account.

\begin{ack}
This work was supported by a Huawei DDMPLab Innovation Research Grant. This project has received funding from the European Union’s Horizon 2020 research and innovation programme under grant agreement No 732204 (Bonseyes).This work is supported by the Swiss State Secretariat for Education, Research and Innovation (SERI) under contract number 16.0159.  The opinions expressed and arguments employed herein do not necessarily reflect the official views of these funding bodies.
\end{ack}


\small
\bibliographystyle{apalike}
\bibliography{main.bib}
\FloatBarrier
\clearpage
\appendix 

\section{Kernels}
\label{sec:appendix_kernels}
\FloatBarrier

\textbf{Polynomial.} This computes a covariance matrix based on the Polynomial kernel between inputs
\begin{equation}\label{eq_kernel_polynomial}
    k^\prime(x, x^{\prime}) =
    (x^{\top} x^{\prime} + c)^{p},
\end{equation}
where $p$ is the degree of the polynomial and $c$ is an offset parameter. We used $p=1$ and $p=2$ in our experiments.

\textbf{Radial Basis Function kernel (RBF).} The RBF is a stationary kernel given by the squared Euclidean distance between the two inputs
\begin{equation}\label{eq_kernel_rbf}
    k^\prime(x, x^{\prime}) =
    \exp \left(- \frac{||x-x^{\prime}||^2}{2l^2} \right),
\end{equation}
where $l$ is a lengthscale parameters learned at training time.

\textbf{Mat\'ern kernel.} This is a stationary kernel which is a generalization of the RBF and the absolute exponential kernel. It is parameterized by a value $\nu>0$, commonly chosen as $\nu=1.5$ (giving once-differentiable functions) or $\nu=2.5$ (giving twice differentiable functions). The kernel is defined as follows:
\begin{equation}
    k^\prime(x, x^{\prime}) = |x-x^{\prime}|^{\nu} K_{\nu}(|x-x^{\prime}|).
\end{equation}\label{eq_kernel_matern}
We used a value of $\nu=2.5$ in our experiments.

\textbf{Spectral mixture kernel.} The spectral mixture kernel was introduced by \citet{wilson2013gaussian} as a powerful stationary kernel for estimating periodic functions. The kernel models a spectral density with a Gaussian mixture
\begin{equation}\label{eq_kernel_spectral}
k^\prime(\tau)=\sum_{q=1}^{Q} w_{q} \prod_{p=1}^{P} \exp \left\{-2 \pi^{2} \tau_{p}^{2} v_{q}^{(p)}\right\} \cos \left(2 \pi \tau_{p} \mu_{q}^{(p)}\right),
\end{equation}
where $\tau=x-x^{\prime}$, $w_q$ are weights that specify the contribution of each mixture component, $\mu_q$ are the component periods, and $v_q$ are lengthscales determining how quickly a component varies with the inputs $x$. We used 4 mixtures in our experiments.

\textbf{Cosine similarity kernel (CosSim).} The cosine similarity kernel consists in taking the product between the unit-normalized input vectors
\begin{equation}
    k^\prime(x, x^{\prime}) =
    \frac{x x^{\prime}}{||x|| \ ||x^{\prime}||}.
\end{equation}\label{eq_kernel_cosine_similarity}
The cosine similarity ranges from -1 (opposite) to 1 (same), with 0 indicating decorrelation (orthogonal). Following the suggestions in \citet{wang2019simpleshot} we experimented with another variant,  meaning centering the input vectors through BatchNorm (BN) statistics \cite{ioffe2015batch} before the normalization (BNCosSim).

\section{Training Details}
\label{appendix:training_details}
\FloatBarrier

\textbf{Datasets.} The CUB dataset \citep{wah2011caltech} consists of  11788 images across 200 classes. We divide the dataset in 100 classes for train, 50 for validation, and 50 for test~\citep{hilliard2018few, chen2019closerfewshot}. The mini-ImageNet dataset \citep{ravi2017optimization} consists of a subset of 100 classes (600 images for each class) taken from the ImageNet dataset~\citep{russakovsky2015imagenet}. We use 64 classes for train, 16 for validation and 20 for test, as is common practice~\citep{ravi2017optimization, chen2019closerfewshot}. The Omniglot dataset \citep{lake2011one} contains 1623 black and white characters taken from 50 different languages. Following standard practice, the number of classes is increased to 6492 by adding examples rotated by 90$^{\circ}$, and we use 4114 for training. The EMNIST dataset~\citep{cohen2017emnist} contains single digits and characters from the English alphabet. We split the 62 classes into 31 for validation and 31 for test.

\textbf{Regression.} In the function prediction experiment, we use the same backbone network described in~\citet{finn2017model}: a two-layer MLP, where each layer has 40 units and ReLU activations. We use the Adam optimizer with learning rate $10^{-3}$ over $5 \times 10^{5}$ training iterations. For regression with feature transfer, a network is trained to predict the output of a function over all tasks, before being fine-tuned on a new task (with 1 or 100 steps of size $10^{-3}$).
For the head pose estimation backbone, we use a three-layer convolutional neural network, each with 36 output channels, stride 2, and dilation 2 to downsample the $100\times100$ input images. We train for 100 steps using the Adam optimizer with learning rate $10^{-3}$. 
 
\begin{figure}[H]
\centerline{
\includegraphics[width=0.6\linewidth]{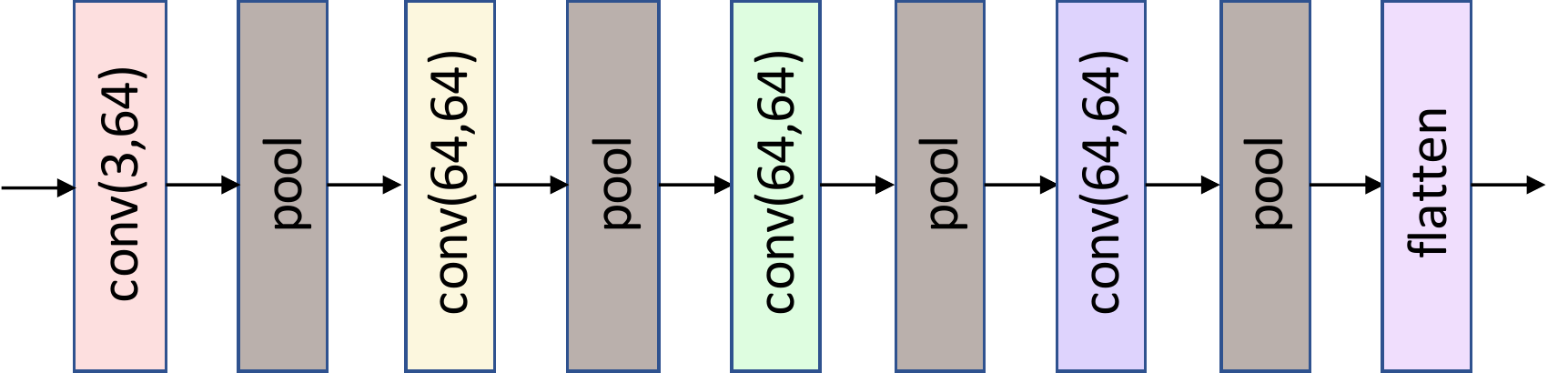}}
\caption{The CNN used as a backbone for classification. It consists of 4 convolutional layers, each consisting of a 2D convolution, a batch-norm layer, and a ReLU non-linearity. The first convolution changes the the number of channels of the input to 64, and the remaining convolutions retain this channel dimension. Each convolutional layer is followed by a max-pooling operation that decreases the spatial resolution of its input by a half. Finally, the output is flattened into a vector when is used as a feature.}
\label{fig_backbone}
\end{figure}

\textbf{Classification.} At training time we apply standard data augmentation (random crop, horizontal flip, and color jitter). The 1-shot training consists of 600 epochs, and 5-shot of 400, for MAML it corresponds to 60000 and 40000 episodes, and for Feature Transfer and Baseline$++$ to 400 and 600 supervised epochs with a mini-batch size of 16. In DKT, the hyperparameters of the kernel are optimized with a learning rate one order of magnitude lower than that used for training the CNN. This helped with convergence. In all experiments we used first-order MAML for memory efficiency. This does not significantly affect results (see~\citealp{chen2019closerfewshot}). In all cases  the validation set has been used to select the training epoch/episode with the best accuracy. In classification and cross-domain experiments, each method uses the same backbone (a four layer CNN), optimizer (Adam), and learning rate ($10^{-3}$). We use shallow backbones because they have been shown to highlight differences between methods~\citep{chen2019closerfewshot}. The CNN used for classification is given in Figure~\ref{fig_backbone}.

\section{Additional Results: Regression Experiments}
\label{sec:appendix_results_regression}

\begin{figure}[H]
\centerline{
\includegraphics[width=1.0\linewidth]{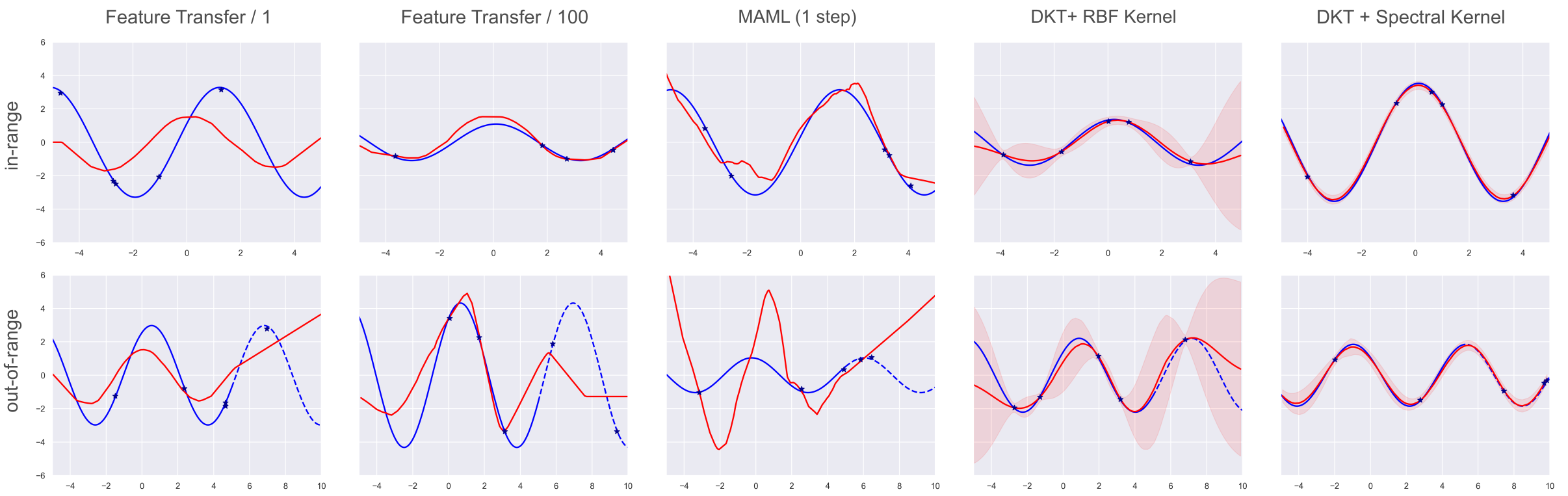}}
\caption{Additional samples for the unknown periodic function prediction experiment. We compare methods for in-range (top row) and out-of-range (bottom row) conditions. The true function is plotted in solid blue, the out-of-range portion in dotted blue, the approximation in red, and the uncertainty is given by a red shadow. The 5 support points (blue stars) are uniformly sampled in the available range.}
\label{fig_plot_periodic_additional}
\end{figure}

\begin{figure}[H]
\centerline{
\includegraphics[width=0.6\linewidth]{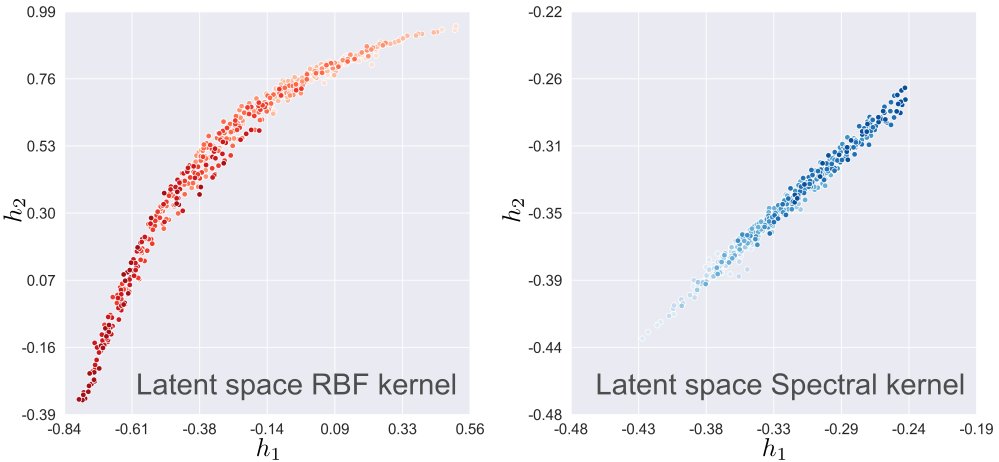}}
\caption{Latent space representation enforced by an RBF (left) and Spectral (right) kernel on the head trajectory experiments.}
\label{fig_latent_space_comparison}
\end{figure}

Here, we provide additional samples of the few-shot regression experiments for a qualitative comparison (Figure~\ref{fig_plot_periodic_additional}). Additionally we compare the latent spaces in the head trajectory estimation experiment. We reduced the number of hidden units to $\mathbf{h}=\{h_1, h_2\}$ and used a hyperbolic tangent activation function (tanh) to project the values to a Cartesian plane with $h_{i} \in [-1,1]$. We then sampled 100 trajectories from the test set and recorded the value of $\mathbf{h}$ for the targets. The resulting plot is shown in Figure~\ref{fig_latent_space_comparison}.  The spectral kernel enforces a more compact manifold, clustering the head poses on a linear gradient based on the value of the target, leading to more accurate predictions.

\section{Additional Results: Classification Experiments}
\label{sec:appendix_results_classification}

\begin{table}[H]
\caption{Average accuracy and standard deviation (percentage) on the few-shot classification setting (5-ways). 
[top] Results reported in recent literature.  
For a fair comparison we selected only those methods that have been trained with a similar backbone and training schedule. [center-bottom] Methods trained from scratch (three runs) with the same backbone (a four layer CNN), optimizer (Adam), and learning rate ($10^{-3}$). Test performed on novel classes with 3000 randomly generated tasks. DKT is competitive across various datasets and conditions. Best results highlighted in bold. $^*$Reported by \citet{jerfel2019reconciling} using a comparable backbone.}
 \begin{adjustbox}{width=1.0\columnwidth,center}
\centering
{
\begin{tabular}{lcccc}
\hline
\textbf{} & \multicolumn{2}{c}{\textbf{CUB}} & \multicolumn{2}{c}{\textbf{mini-ImageNet}} \\
\small{\textbf{Method}} & \textbf{1-shot} & \textbf{5-shot} & \textbf{1-shot} & \textbf{5-shot}\\
\hline
\small{\textbf{ML-LSTM}} \citep{ravi2017optimization} & -- & --  & 43.44 $\pm$ \small{0.77} & 60.60 $\pm$ \small{0.71}\\
\small{\textbf{SNAIL}} \citep{mishra2018simple} & -- & --  & 45.10 & 55.20\\
\small{\textbf{iMAML-HF}} \citep{rajeswaran2019meta} & -- & --  & 49.30 $\pm$ \small{1.88} & --\\
\small{\textbf{LLAMA}} \citep{grant2018recasting} & -- & --  & 49.40 $\pm$ \small{1.83} & --\\
\small{\textbf{VERSA}} \citep{gordon2019meta}$^*$ & -- & --  & 48.53 $\pm$ \small{1.84} & --\\
\small{\textbf{Amortized VI}} \citep{gordon2019meta} &  -- & --  & 44.13 $\pm$ \small{1.78} & 55.68 $\pm$ \small{0.91}\\
\small{\textbf{Meta-Mixture}} \citep{jerfel2019reconciling} &  -- & --  & 49.60 $\pm$ \small{1.50} & 64.60 $\pm$ \small{0.92}\\
\small{\textbf{SimpleShot}} \citep{wang2019simpleshot} &  -- & --  & 49.69 $\pm$ \small{0.19} & \textbf{66.92 $\pm$ \small{0.17}}\\
\hline
\small{\textbf{Feature Transfer}}
& 46.19 $\pm$ \small{0.64} & 68.40 $\pm$ \small{0.79} & 39.51 $\pm$ \small{0.23} & 60.51 $\pm$ \small{0.55}\\
\small{\textbf{Baseline$++$}} \citep{chen2019closerfewshot} & 61.75 $\pm$ \small{0.95} & \textbf{78.51 $\pm$ \small{0.59}} & 47.15 $\pm$ \small{0.49} & 66.18 $\pm$ \small{0.18}\\
\small{\textbf{MatchingNet}} \citep{vinyals2016matching} & 60.19 $\pm$ \small{1.02} & 75.11 $\pm$ \small{0.35} & 48.25 $\pm$ \small{0.65} & 62.71 $\pm$ \small{0.44} \\
\small{\textbf{ProtoNet}} \citep{snell2017prototypical} & 52.52 $\pm$ \small{1.90} & 75.93 $\pm$ \small{0.46} &44.19 $\pm$ \small{1.30} & 64.07 $\pm$ \small{0.65} \\
\small{\textbf{MAML}}  \citep{finn2017model} & 56.11 $\pm$ \small{0.69} & 74.84 $\pm$ \small{0.62}  &45.39 $\pm$ \small{0.49} & 61.58 $\pm$ \small{0.53} \\
\small{\textbf{RelationNet}} \citep{sung2018learning} & 62.52 $\pm$ \small{0.34} & 78.22 $\pm$ \small{0.07}  & 48.76 $\pm$ \small{0.17} & 64.20 $\pm$ \small{0.28}\\
\hline
\small{\textbf{DKT + CosSim}} (ours) & \textbf{63.37 $\pm$ \small{0.19}} & 77.73 $\pm$ \small{0.26} & 48.64 $\pm$ \small{0.45} & 62.85 $\pm$ \small{0.37} \\
\small{\textbf{DKT + BNCosSim}} (ours) & 62.96 $\pm$ \small{0.62} & 77.76 $\pm$ \small{0.62} & \textbf{49.73 $\pm$ \small{0.07}} & 64.00 $\pm$ \small{0.09} \\
\hline
\end{tabular}
}
\label{tab_classification_accuracy}
 \end{adjustbox}
\end{table}

\textbf{Kernel comparison.} In Table~\ref{tab_classification_kernel_comparison} we show a comparison between different kernels (linear, RBF, Mat\'ern, Polynomial $p=1$ and $p=2$, CosSim, BNCosSim) trained on CUB and mini-ImageNet. In this setting using a BNCosSim kernel gives a large advantage in almost all conditions. This result is in line with the findings of \citet{wang2019simpleshot}, who showed how centering and unit normalizing the features considerably improve the performance in classification tasks. The overall performance of CosSim and BNCosSim is also in accordance with the findings of \citet{chen2019closerfewshot} and their implementation of Baseline$++$, an effective feature transfer method based on the cosine distance. Further investigations are necessary in this direction to understand the reason why cosine metrics and normalization are so important in few-shot learning.

\begin{table}[H]
\caption{Average accuracy and standard deviation (percentage) on the few-shot classification setting (5-ways) for different kernels. Methods trained from scratch (three runs) with the same backbone (a four layer CNN), optimizer (Adam), and learning rate ($10^{-3}$). Test performed on novel classes with 3000 randomly generated tasks.}
\centering
\begin{tabular}{lcccc}
\hline
\textbf{} & \multicolumn{2}{c}{\textbf{CUB}} & \multicolumn{2}{c}{\textbf{mini-ImageNet}} \\
\small{\textbf{Kernel}} & \textbf{1-shot} & \textbf{5-shot} & \textbf{1-shot} & \textbf{5-shot}\\
\hline
\small{\textbf{Linear}} & 60.23 $\pm$ \small{0.76} & 74.74 $\pm$ \small{0.22} & 48.44 $\pm$ \small{0.36} & 62.88 $\pm$ \small{0.46} \\
\small{\textbf{RBF}} & 55.34 $\pm$ \small{2.56} & 73.20 $\pm$ \small{1.41} & 45.92 $\pm$ \small{1.08} & 61.42 $\pm$ \small{0.74} \\
\small{\textbf{Mat\'ern}} & 58.20 $\pm$ \small{0.63} & 73.21 $\pm$ \small{1.30} & 47.65 $\pm$ \small{0.85} & 62.59 $\pm$ \small{0.12} \\
\small{\textbf{Polynomial ($p=1$)}} & 59.54 $\pm$ \small{1.10} & 74.51 $\pm$ \small{0.98} & 47.78 $\pm$ \small{0.60} & 62.54 $\pm$ \small{0.96} \\
\small{\textbf{Polynomial ($p=2$)}} & 5718 $\pm$ \small{0.40} & 71.14 $\pm$ \small{0.58} & 46.36 $\pm$ \small{0.34} & 60.26 $\pm$ \small{0.40} \\
\small{\textbf{CosSim}} & \textbf{63.37 $\pm$ \small{0.19}} & 77.73 $\pm$ \small{0.26} & 48.64 $\pm$ \small{0.45} & 62.85 $\pm$ \small{0.37} \\
\small{\textbf{BNCosSim}} & 62.96 $\pm$ \small{0.62} & \textbf{77.76 $\pm$ \small{0.62}} & \textbf{49.73 $\pm$ \small{0.07}} & \textbf{64.00 $\pm$ \small{0.09}} \\
\hline
\end{tabular}
\label{tab_classification_kernel_comparison}
\end{table}

\begin{table}[H]
\caption{Average accuracy and standard deviation (percentage) over three runs on 1-shot and 5-shot classification (5-ways), for different backbones in the CUB dataset. We use the same setup as in the classification setting. The results for the ResNet are the ones reported in \citet{chen2019closerfewshot}. DKT has the best score in 1-shot Conv-4, and 5-shot ResNet, while being competitive in the other conditions. Best results highlighted in bold.}
\centering
\begin{tabular}{lcccc}
\hline
\textbf{} & \multicolumn{2}{c}{\textbf{Conv-4}} & \multicolumn{2}{c}{\textbf{ResNet-10}} \\
\small{\textbf{Method}} & \textbf{1-shot}& \textbf{5-shot} & \textbf{1-shot} & \textbf{5-shot} \\ 
\hline
\small{\textbf{Feature Transfer}}
& 46.19 $\pm$ \small{0.64} & 68.40 $\pm$ \small{0.79} & 63.64 $\pm$ \small{0.91} & 81.27 $\pm$ \small{0.57}\\
\small{\textbf{Baseline$++$}} \citep{chen2019closerfewshot} & 61.75 $\pm$ \small{0.95} & \textbf{78.51 $\pm$ \small{0.59}} & 69.55 $\pm$ \small{0.89} & 85.17 $\pm$ \small{0.50}\\
\small{\textbf{MatchingNet}} \citep{vinyals2016matching} & 60.19 $\pm$ \small{1.02} & 75.11 $\pm$ \small{0.35} & 71.29 $\pm$ \small{0.87} & 83.47 $\pm$ \small{0.58}\\
\small{\textbf{ProtoNet}} \citep{snell2017prototypical} & 52.52 $\pm$ \small{1.90} & 75.93 $\pm$ \small{0.46} & \textbf{73.22 $\pm$ \small{0.92}} & 85.01 $\pm$ \small{0.52}\\
\small{\textbf{MAML}}  \citep{finn2017model} & 56.11 $\pm$ \small{0.69} & 74.84 $\pm$ \small{0.62}  & 70.32 $\pm$ \small{0.99} & 80.93 $\pm$ \small{0.71}\\
\small{\textbf{RelationNet}} \citep{sung2018learning} & 62.52 $\pm$ \small{0.34} & 78.22 $\pm$ \small{0.07}  & 70.47 $\pm$ \small{0.99} & 83.70 $\pm$ \small{0.55}\\
\hline
\small{\textbf{DKT + CosSim}} (ours) & \textbf{63.37 $\pm$ \small{0.19}} & 77.73 $\pm$ \small{0.26} & 70.81 $\pm$ \small{0.52} & 83.26 $\pm$ \small{0.50}\\
\small{\textbf{DKT + BNCosSim}} (ours) & 62.96 $\pm$ \small{0.62} & 77.76 $\pm$ \small{0.62} & 72.27 $\pm$ \small{0.30} & \textbf{85.64 $\pm$ \small{0.29}}\\
\hline
\end{tabular}
\label{tab_backbone_accuracy}
\end{table}

\begin{table}[H]
\caption{Average \emph{Expected Calibration Error} (\emph{ECE}, \citealt{guo2017calibration}) with standard deviation (percentage) over three runs on 1-shot and 5-shot classification (5-ways) in the CUB dataset. The lower the better. For the training phase we used the same setup as in the classification experiments. In the evaluation phase, the temperature of all models has been calibrated on 3000 randomly generated tasks, then each method has been evaluated on a separate set of 3000 randomly generated test tasks. DKT has the third lowest error in 1-shot, and the second lowest error in 5-shot. Best results highlighted in bold.}
\centering
\begin{tabular}{lcc}
\hline
\small{\textbf{Method}} & \textbf{1-shot}& \textbf{5-shot} \\ 
\hline
\small{\textbf{Feature Transfer}} & 12.57 $\pm$ \small{0.23} & 18.43 $\pm$ \small{0.16} \\
\small{\textbf{Baseline$++$}} \citep{chen2019closerfewshot} & 4.91 $\pm$ \small{0.81} & 2.04 $\pm$ \small{0.67} \\
\small{\textbf{MatchingNet}} \citep{vinyals2016matching} & 3.11 $\pm$ \small{0.39} & 2.23 $\pm$ \small{0.25}  \\
\small{\textbf{ProtoNet}} \citep{snell2017prototypical} & \textbf{1.07 $\pm$ \small{0.15}} & \textbf{0.93 $\pm$ \small{0.16}}  \\
\small{\textbf{MAML}}  \citep{finn2017model} & 1.14 $\pm$ \small{0.22} & 2.47 $\pm$ \small{0.07}  \\
\small{\textbf{RelationNet}} \citep{sung2018learning} & 4.13 $\pm$ \small{1.72} & 2.80 $\pm$ \small{0.63}  \\
\hline
\small{\textbf{DKT + BNCosSim}} (ours) & 2.62 $\pm$ \small{0.19} & 1.15 $\pm$ \small{0.21}  \\
\hline
\end{tabular}
\label{tab_uncertainty_ece}
\end{table}

\section{Additional Results: Cross-Domain Experiments}
\label{sec:appendix_results_crossdomain}

\begin{table}[H]
\caption{Average accuracy and standard deviation (percentage) over three runs on the cross-domain setting (5-ways). We use the same setup as in the classification setting. The proposed method (DKT) has the best score on most conditions. Best results highlighted in bold.}
\centering
\begin{tabular}{lcccc}
\hline
\textbf{} & \multicolumn{2}{c}{\textbf{Omniglot}$\rightarrow$\textbf{EMNIST}} & \multicolumn{2}{c}{\textbf{mini-ImageNet}$\rightarrow$\textbf{CUB}} \\
\small{\textbf{Method}} & \textbf{1-shot}& \textbf{5-shot} & \textbf{1-shot} & \textbf{5-shot} \\ 
\hline
\small{\textbf{Feature Transfer}} & 64.22 $\pm$ \small{1.24} & 86.10 $\pm$ \small{0.84} & 32.77 $\pm$ \small{0.35} & 50.34 $\pm$ \small{0.27}\\
\small{\textbf{Baseline$++$}} \citep{chen2019closerfewshot} & 56.84 $\pm$ \small{0.91} & 80.01 $\pm$ \small{0.92} & 39.19 $\pm$ \small{0.12} & \textbf{57.31 $\pm$ \small{0.11}}\\
\small{\textbf{MatchingNet}} \citep{vinyals2016matching}  & 75.01 $\pm$ \small{2.09} & 87.41 $\pm$ \small{1.79} & 36.98 $\pm$ \small{0.06} & 50.72 $\pm$ \small{0.36} \\
\small{\textbf{ProtoNet}} \citep{snell2017prototypical} & 72.04 $\pm$ \small{0.82} & 87.22 $\pm$ \small{1.01} & 33.27 $\pm$ \small{1.09} & 52.16 $\pm$ \small{0.17} \\
\small{\textbf{MAML}} \citep{finn2017model} & 72.68 $\pm$ \small{1.85} & 83.54 $\pm$ \small{1.79}  & 34.01 $\pm$ \small{1.25} &48.83 $\pm$ \small{0.62} \\
\small{\textbf{RelationNet}} \citep{sung2018learning} & 75.62 $\pm$ \small{1.00} & 87.84 $\pm$ \small{0.27}  & 37.13 $\pm$ \small{0.20} & 51.76 $\pm$ \small{1.48}\\
\hline
\small{\textbf{DKT + Linear}} (ours) & \textbf{75.97 $\pm$ \small{0.70}} & 89.51 $\pm$ \small{0.44} & 38.72 $\pm$ \small{0.42} & 54.20 $\pm$ \small{0.37}\\
\small{\textbf{DKT + CosSim}} (ours) & 73.06 $\pm$ \small{2.36} & 88.10 $\pm$ \small{0.78} & \textbf{40.22 $\pm$ \small{0.54}} & 55.65 $\pm$ \small{0.05} \\
\small{\textbf{DKT + BNCosSim}} (ours) & 75.40 $\pm$ \small{1.10} & \textbf{90.30 $\pm$ \small{0.49}} & 40.14 $\pm$ \small{0.18} & 56.40 $\pm$ \small{1.34} \\
\hline
\end{tabular}
\label{tab_crossdomain_accuracy}
\end{table}

\textbf{Kernel comparison.} In Table~\ref{tab_crossdomain_kernel_comparison} we show a comparison between different kernels (linear, RBF, Mat\'ern, Polynomial $p=1$ and $p=2$, CosSim, BNCosSim) trained on Omniglot$\rightarrow$EMNIST and mini-ImageNet$\rightarrow$CUB. Overall using a BNCosSim kernel still gives an advantage in almost all conditions, showing stable results. The best accuracy is achieved using more specialized kernels, however they often reach peak performance in specific conditions while underperforming in others.

\begin{table}[H]
\caption{Average accuracy and standard deviation (percentage) over three runs on the cross-domain setting (5-ways) for different kernels. We use the same setup as in the classification setting.}
\centering
\begin{tabular}{lcccc}
\hline
\textbf{} & \multicolumn{2}{c}{\textbf{Omniglot}$\rightarrow$\textbf{EMNIST}} & \multicolumn{2}{c}{\textbf{mini-ImageNet}$\rightarrow$\textbf{CUB}} \\
\small{\textbf{Kernel}} & \textbf{1-shot}& \textbf{5-shot} & \textbf{1-shot} & \textbf{5-shot} \\ 
\hline
\small{\textbf{Linear}} & \textbf{75.97 $\pm$ \small{0.70}} & 89.51 $\pm$ \small{0.44} & 38.72 $\pm$ \small{0.42} & 54.20 $\pm$ \small{0.37}\\
\small{\textbf{RBF}} & 74.46 $\pm$ \small{0.41} & 88.38 $\pm$ \small{0.53} & 36.22 $\pm$ \small{0.40} & 51.30 $\pm$ \small{0.52} \\
\small{\textbf{Mat\'ern}} & 75.46 $\pm$ \small{0.20} & 88.04 $\pm$ \small{1.81} & 36.98 $\pm$ \small{0.41} & 51.35 $\pm$ \small{0.16} \\
\small{\textbf{Polynomial ($p=1$)}} & 74.33 $\pm$ \small{0.67} & \textbf{90.72 $\pm$ \small{0.47}} & 38.24 $\pm$ \small{0.30} &  54.11 $\pm$ \small{0.40} \\
\small{\textbf{Polynomial ($p=2$)}} & 75.58 $\pm$ \small{1.18} & 88.06 $\pm$ \small{0.70} & 36.83 $\pm$ \small{0.46} &  51.92 $\pm$ \small{0.87} \\
\small{\textbf{CosSim}} & 73.06 $\pm$ \small{2.36} & 88.10 $\pm$ \small{0.78} & \textbf{40.22 $\pm$ \small{0.54}} & 55.65 $\pm$ \small{0.05} \\
\small{\textbf{BNCosSim}} & 75.40 $\pm$ \small{1.10} & 90.30 $\pm$ \small{0.49} & 40.14 $\pm$ \small{0.18} & \textbf{56.40 $\pm$ \small{1.34}} \\
\hline
\end{tabular}
\label{tab_crossdomain_kernel_comparison}
\end{table}

\end{document}